\documentclass[sponsored]{acmsiggraph}

\usepackage[utf8]{inputenc}
\usepackage{makeidx}  
\usepackage{graphics}
\usepackage{caption}
\usepackage{subcaption}
\usepackage{here}
\usepackage{enumitem}
\usepackage{color, colortbl}
\usepackage{amssymb}
\usepackage{amsfonts}
\usepackage[]{algorithm2e}
\usepackage[T1]{fontenc}
\usepackage[table]{xcolor}
\usepackage{booktabs}

\definecolor{lightgray}{gray}{0.9}

\title{ElSe: Ellipse Selection for Robust Pupil Detection in Real-World Environments}

\author{Wolfgang Fuhl\thanks{wolfgang.fuhl@uni-tuebingen.de}\\ Perception Engineering\\ University of Tuebingen %
\and Thiago C. Santini\thanks{thiago.santini@uni-tuebingen.de}\\ Perception Engineering\\ University of Tuebingen %
\and Thomas Kübler\thanks{thomas.kuebler@uni-tuebingen.de}\\ Perception Engineering\\ University of Tuebingen %
\and Enkelejda Kasneci\thanks{enkelejda.kasneci@uni-tuebingen.de}\\ Perception Engineering\\ University of Tuebingen}

\pdfauthor{Wolfgang Fuhl, Thiago C. Santini, Thomas Kuebler, Enkelejda Kasneci}


\begin{document}

\maketitle

\begin{abstract}
	
Fast and robust pupil detection is an essential prerequisite for video-based eye-tracking in real-world settings. Several algorithms for image-based pupil detection have been proposed, their applicability is mostly limited to laboratory conditions. In real-world scenarios, automated pupil detection has to face various challenges, such as illumination changes, reflections (on glasses), make-up, non-centered eye recording, and physiological eye characteristics. We propose \emph{ElSe}, a novel algorithm based on ellipse evaluation of a filtered edge image. We aim at a robust, resource-saving approach that can be integrated in embedded architectures e.g. driving. The proposed algorithm was evaluated against four state-of-the-art methods on over 93,000 hand-labeled images from which 55,000 are new images contributed by this work. On average, the proposed method achieved a 14.53\% improvement on the detection
rate relative to the best state-of-the-art performer. download:\textit{ftp://emmapupildata@messor.informatik.uni-tuebingen.de} (password:\textit{eyedata}).

\end{abstract}

\keywordlist

\copyrightspace

\section{Introduction}
Understanding processes underlying human visual perception has been in the focus of various research in the fields of medicine, psychology, advertisement, autonomous cars, application control, and
many more.
Over the last years head-mounted, mobile eye trackers enabled the measurement
and investigation of the human viewing behavior in real-world and dynamic tasks.
Such eye trackers map the gaze point of the scene based on the center of the automatically detected pupil in the eye images. 

While tracking can be accomplished successfully under laboratory conditions, many studies report the occurrence of difficulties when eye trackers are employed in natural environments, such as driving~\cite{kasneci13,liu2002real} or
shopping~\cite{kasneci2014homonymous,supermarkt-glaucoma}. The main source of
noise in such experimental tasks is a non-robust pupil signal that is mainly
related to challenges in the image-based detection of the pupil.
\cite{schnipke2000trials} summarized a variety of difficulties occurring when
using eye trackers, such as changing illumination, motion blur, recording errors
and eyelashes covering the pupil. Rapidly changing illumination conditions arise
primarily in tasks where the subject is moving fast, e.g., while driving, or
rotates relative to unequally distributed light sources. Furthermore, in case
the subject is wearing eye glasses or contact lenses, further reflections may
occur. Another issue arises due to the off-axial position of eye camera in
head-mounted eye trackers. Therefore, studies based on eye tracking outside of
laboratory constantly report low pupil detection rates
\cite{kasneci2014driving,liu2002real,trosterer2014eye}. As a consequence, the
data collected in such studies has to be manually post-processed, which is a
laborious and time-consuming procedure.
Furthermore, this post-processing is impossible for real-time applications that rely on the pupil monitoring (e.g., driving or surgery assistance). Such real-time applications also impose harsh constraints on the algorithm, making the use of  computer-intensive methods impracticable and leading to the prevalence of threshold-based methods.
\\
Several algorithms address image-based pupil detection under laboratory
conditions. For example, \cite{goni2004robust} use a histogram-based threshold
calculation on bright pupils. A similar approach was introduced
by~\cite{keil2010real}, where a corneal reflection detection was performed on
top of a histogram-based method. Such algorithms can be applied to eye images
captured under infrared light as in~\cite{lin2010robust}
and~\cite{long2007high}. In \cite{long2007high} and \cite{perez2003precise}, the
center of the pupil is estimated based on a threshold and a center of mass
calculation. Another threshold-based approach was presented
in~\cite{zhu1999robust}, where the pupil is detected based on the calculation of
the curvature of the threshold border. A similar approach is also used by the
recently published algorithm SET~\cite{javadi2015set}, which first extracts
pupil pixels based on a luminance threshold. Afterwards, the shape of the
thresholded area is extracted and compared against a sine curve. An isophotes
curvature-based approach is presented by~\cite{roberto2012accurate} using the
maximum isocenter as pupil center estimation. Despite recent developments the
most popular algorithm in this realm is probably Starburst, introduced
by~\cite{li2005starburst}. Starburst sends out rays in multiple directions and
collects all positions where the difference of consecutive pixels is higher
than a threshold. The mean position is calculated and this step is repeated
until convergence. \cite{swirski2012robust} proposed an algorithm starting with
a coarse positioning using Haar-like features. The intensity histogram of the
coarse position is clustered using k-means followed by a modified RANSAC ellipse
fit.
\cite{fuhl2015excuse}~proposed ExCuSe, which was designed with the
aforementioned challenges that arise from real-world scenarios in mind.  Based
on an intensity histogram analysis the algorithm decides whether the input image
has reflections or not. On images with reflections the edge image is filtered
and the best curve is selected. Otherwise it starts with a coarse position
followed by a refinement.\\
In this paper, we present a novel algorithm for pupil detection named
\textit{Ellipse Selector} (\textit{ElSe} for short) based on edge filtering,
ellipse evaluation, and pupil validation. We evaluated ElSe on over 94,000
images collected during eye-tracking experiments in real-world scenarios. ElSe
proved high detection rates, robustness, and a fast runtime in comparison to
state-of-the-art algorithms ExCuSe~\cite{fuhl2015excuse},
SET~\cite{javadi2015set}, Starburst~\cite{li2005starburst}, and~\cite{swirski2012robust}. As an additional contribution, both
data set and the annotated pupil centers are openly accessible to support
further research.

\section{Method}

\begin{figure}[h]
	\centering
	\includegraphics[width=0.4\textwidth,natwidth=1297,natheight=1010]{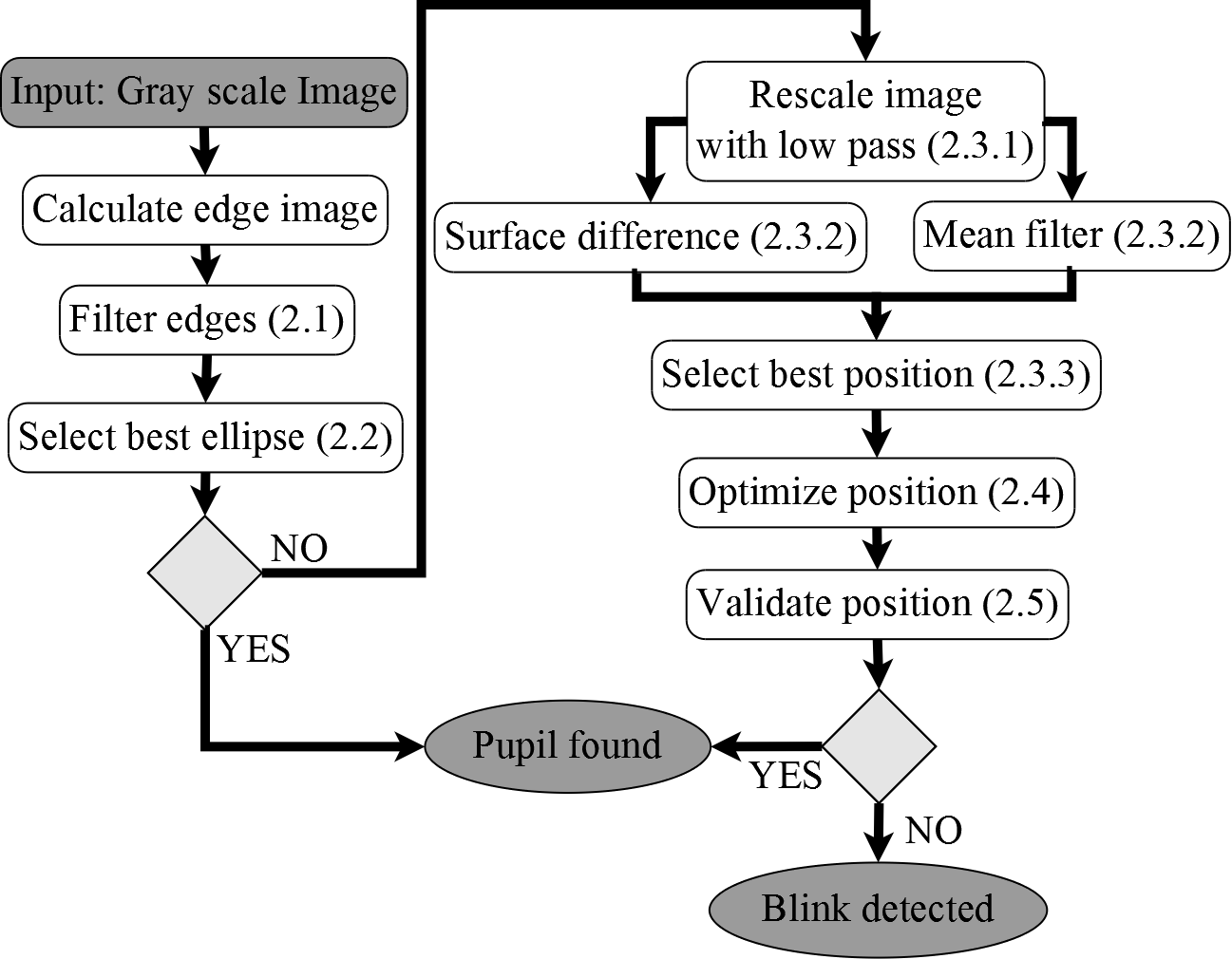}
	\caption{Flowchart of the proposed algorithm. Light gray boxes represent decisions, dark gray ellipses termination points, and white boxes represent processing steps.}
	\label{fig:ablauf}
\end{figure}
ElSe operates on gray scale images. To reduce the effect of eyeglass
frames, 10\% of the border area of the image is excluded from processing. After normalization, a Canny edge
filter is applied to the image (Figure~\ref{fig:ablauf}).
In the next algorithmic step (Step 2.1 in Figure~\ref{fig:ablauf}), edge connections that could impair the surrounding
edge of the pupil are removed. Afterwards, in Step 2.2, connected edges are
collected and evaluated based on straightness, inner intensity value, elliptic
properties, the possibility to fit an ellipse to it, and a pupil plausibility
check. If a valid ellipse describing the pupil is found, it is returned as the
result. In case no ellipse is found (e.g.,
when the edge filtering does not result in suitable edges), a second analysis is conducted. To speed up the convolution with the surface difference (Step 2.3.2) and mean filter (Step 2.3.2), the image is downscaled (Step 2.3.1). This operation is performed by calculating a histogram for all pixels from the large image influencing the pixel in the downscaled image. In each histogram, the mean of all intensity values up to the mean of the histogram is calculated and used as a value for the pixel in the downscaled image. After applying the surface difference and mean filter to the rescaled image, the best position is selected (Step 2.3.3) by multiplying the result of both filters and selecting the maximum position. Choosing a pixel position in the downscaled image leads to a distance error of the pupil center in the full scale image. Therefore, the position has to be optimized on the full scale image (Step 2.4) based on an analysis of the surrounding pixels of the
chosen position.  
In the following sections, each of the above mentioned processing steps is described in detail.
\subsection{Filter edges}
\begin{figure}[h]
	\centering
	\begin{subfigure}[b]{0.33\marginparwidth}
		\centering
		\includegraphics[height=0.075\textheight,natwidth=182,natheight=462]{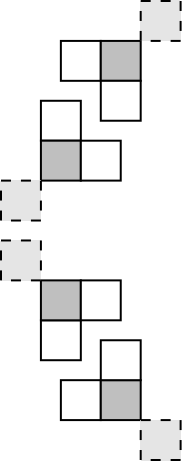}
		\caption{}
		\label{fig:morphA}
	\end{subfigure}
	\begin{subfigure}[b]{0.39\marginparwidth}
		\centering
		\includegraphics[height=0.075\textheight,natwidth=122,natheight=422]{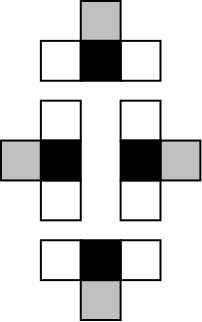}
		\caption{}
		\label{fig:morphB}
	\end{subfigure}
	\begin{subfigure}[b]{0.39\marginparwidth}
		\centering
		\includegraphics[height=0.075\textheight,natwidth=201,natheight=582]{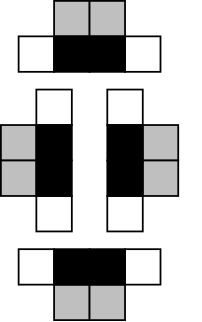}
		\caption{}
		\label{fig:morphC}
	\end{subfigure}
	\begin{subfigure}[b]{0.33\marginparwidth}
		\centering
		\includegraphics[height=0.075\textheight,natwidth=202,natheight=322]{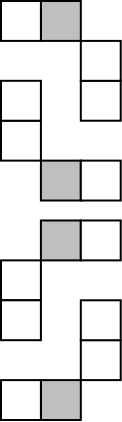}
		\caption{}
		\label{fig:morphD}
	\end{subfigure}
	\begin{subfigure}[b]{0.33\marginparwidth}
		\centering
		\includegraphics[height=0.075\textheight,natwidth=202,natheight=322]{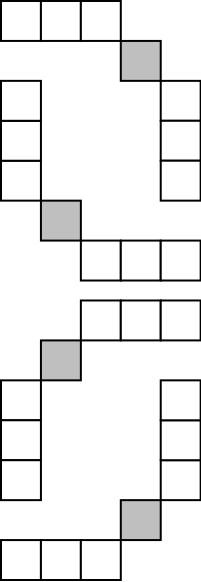}
		\caption{}
		\label{fig:morphE}
	\end{subfigure}
	\begin{subfigure}[b]{0.33\marginparwidth}
		\centering
		\includegraphics[height=0.075\textheight,natwidth=282,natheight=482]{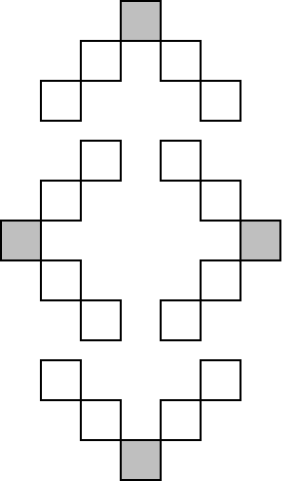}
		\caption{}
		\label{fig:morphF}
	\end{subfigure}
	\begin{subfigure}[b]{0.33\marginparwidth}
		\centering
		\includegraphics[height=0.075\textheight,natwidth=282,natheight=522]{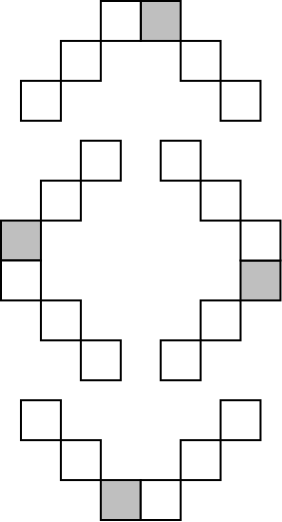}
		\caption{}
		\label{fig:morphG}
	\end{subfigure}
	\caption{Morphologic patterns for edge manipulation. White and
		dark gray boxes represent pixels that have to remain edge pixels. Light gray boxes with dashed borders (\subref{fig:morphA}) represent pixels that have to be removed. If the pattern matches a segment in the edge image, pixels under dark gray boxes are removed from, and pixels under black boxes are added to the edge image.  The pattern in (\subref{fig:morphA}) thins lines, whereas patterns (\subref{fig:morphB}) and (\subref{fig:morphC}) straightens lines. The patterns (\subref{fig:morphD}), (\subref{fig:morphE}), (\subref{fig:morphF}), and
		(\subref{fig:morphG}) are applied to break up orthogonal connections.
	}
	\label{fig:morph}
\end{figure}
Edges are split up at positions that do not occur in an ellipse, e.g., orthogonal connectors and edge points with more than two neighbors. Additionally, edges are thinned and straightened in order to improve the breaking procedure based on two approaches (morphologic and algorithmic). Both approaches lead to comparable results. In the provided implementation, ElSe uses the morphologic approach because it requires less computational
power.

\subsubsection{Morphologic approach}
\begin{figure}[h]
	\centering
	\begin{subfigure}[b]{0.6\marginparwidth}
		\centering
		\includegraphics[width=0.99\textwidth,natwidth=100,natheight=100]{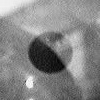}
		\caption{}
		\label{fig:morph_resultA}
	\end{subfigure}
	\begin{subfigure}[b]{0.6\marginparwidth}
		\centering
		\includegraphics[width=0.99\textwidth,natwidth=100,natheight=100]{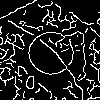}
		\caption{}
		\label{fig:morph_resultB}
	\end{subfigure}
	\begin{subfigure}[b]{0.6\marginparwidth}
		\centering
		\includegraphics[width=0.99\textwidth,natwidth=100,natheight=100]{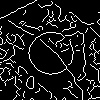}
		\caption{}
		\label{fig:morph_resultC}
	\end{subfigure}
	
	\begin{subfigure}[b]{0.6\marginparwidth}
		\centering
		\includegraphics[width=0.99\textwidth,natwidth=102,natheight=102]{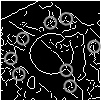}
		\caption{}
		\label{fig:morph_resultD}
	\end{subfigure}
	\begin{subfigure}[b]{0.6\marginparwidth}
		\centering
		\includegraphics[width=0.99\textwidth,natwidth=102,natheight=102]{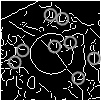}
		\caption{}
		\label{fig:morph_resultE}
	\end{subfigure}
	\begin{subfigure}[b]{0.6\marginparwidth}
		\centering
		\includegraphics[width=0.99\textwidth,natwidth=102,natheight=102]{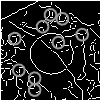}
		\caption{}
		\label{fig:morph_resultF}
	\end{subfigure}
	\caption{
		(\subref{fig:morph_resultA}) input image,
		(\subref{fig:morph_resultB}) edge filtered results,
		(\subref{fig:morph_resultC}) edges after thinning using the morphologic pattern from Figure~\ref{fig:morphA}.
		(\subref{fig:morph_resultD}) remaining edges after deleting all edges
		with too many neighbors.
		(\subref{fig:morph_resultE}) result of edge straightening by applying
		the two morphologic operations shown in Figures~\ref{fig:morphB} and
		\ref{fig:morphC}. (\subref{fig:morph_resultF}) result after deleting edge pixels that connect orthogonal by means of the morphologic patterns shown in
		Figures~\ref{fig:morphD}, \ref{fig:morphE}, \ref{fig:morphF}, and
		\ref{fig:morphG}.
	}
	\label{fig:morph_result}
\end{figure}
The employed morphologic operations in Figures~\ref{fig:morphB}, \ref{fig:morphC}, \ref{fig:morphD}, and \ref{fig:morphE} are similar to those introduced in
ExCuSe~\cite{fuhl2015excuse}. However, in contrast to it, no
preprocessing based on deletion of edges with low angle is performed.
Furthermore, we introduce a stable thinning procedure (Figure~\ref{fig:morphA}) and deletion of edges with too many neighbors.
The morphologic processing starts with edge-thinning using the pattern shown in
Figure~\ref{fig:morphA}. Figure~\ref{fig:morph_resultC} presents the result of
thinning applied on the Canny edge image from Figure~\ref{fig:morph_resultB}.
Afterwards, the direct neighborhood of each edge pixel is summed up. If this
neighborhood is $>2$, the edge pixel is deleted because it has joined more than
two lines. Applied to the result from the thinning step,
Figure~\ref{fig:morph_resultD} shows the remaining edge pixels.
Next, a refinement step is performed by applying the straightening
patterns in Figure~\ref{fig:morphB} and \ref{fig:morphC}, yielding the
edges in Figure~\ref{fig:morph_resultE}.
Then, the patterns shown in Figure~\ref{fig:morphD},
\ref{fig:morphE}, \ref{fig:morphF}, and \ref{fig:morphG} are applied; as a
result, the orthogonal connections in consecutive edge pixels are separated by
deleting the connecting pixel, resulting in Figure~\ref{fig:morph_resultF}.
\begin{figure}[h]
	\centering
	\begin{subfigure}[b]{1.2\marginparwidth}
		\centering
		\includegraphics[width=1.0\textwidth,natwidth=785,natheight=317]{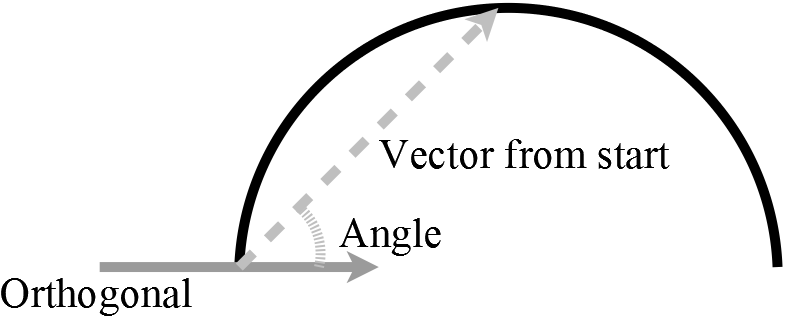}
		\caption{}
		\label{fig:algo_break_upA}
	\end{subfigure}
	\hfil
	\begin{subfigure}[b]{0.7\marginparwidth}
		\centering
		\includegraphics[width=1.0\textwidth,natwidth=956,natheight=462]{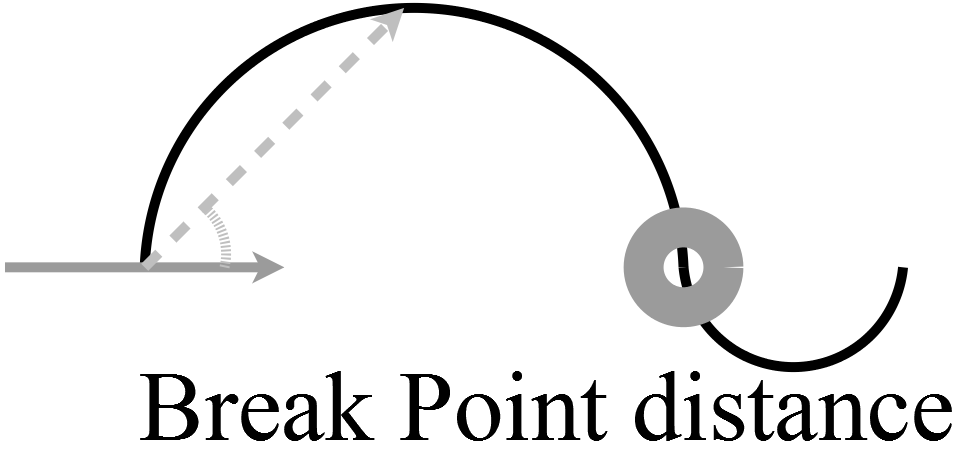}
		\caption{}
		\label{fig:algo_break_upB}
	\end{subfigure}
	\begin{subfigure}[b]{0.7\marginparwidth}
		\centering
		\includegraphics[width=1.0\textwidth,natwidth=1026,natheight=322]{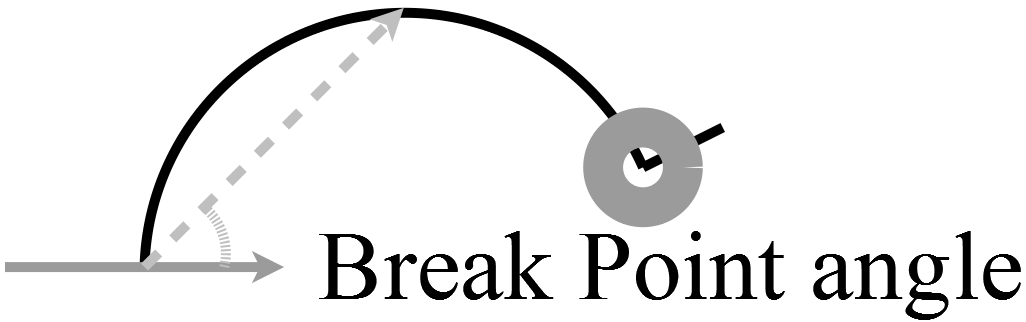}
		\caption{}
		\label{fig:algo_break_upC}
	\end{subfigure}
	\caption{ In (\subref{fig:algo_break_upA}), the gray arrow is the
		calculated orthogonality, whereas the dashed gray arrow is the vector
		between the starting and the current point. The black line
		represents the processed edge. As the gray dashed arrow moves along the
		edge, the angle to the orthogonal decreases, while the length of the vector increases.
		(\subref{fig:algo_break_upB}) distance
		breaking and (\subref{fig:algo_break_upC}) angle breaking condition is triggered.
	}
	\label{fig:algo_break_up}
\end{figure}

\subsubsection{Algorithmic approach}
\begin{figure}[h]
	\centering
	\includegraphics[width=0.45\textwidth,natwidth=726,natheight=668]{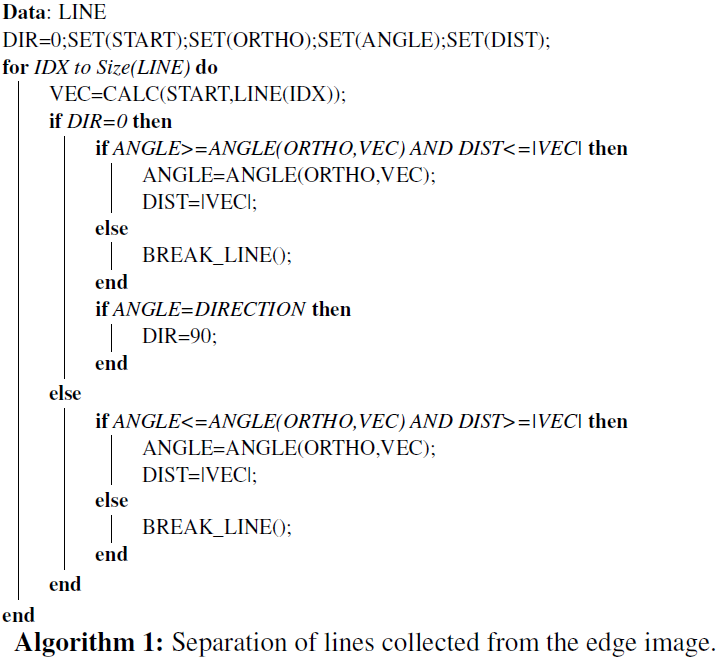}
	\label{alg:line_break}
\end{figure}

The algorithmic approach to filtering the edge
image is based on the idea of breaking up lines at positions where the line
course cannot belong to a common ellipse. Prerequisites here are edge-thinning,
breaking up lines with too many neighbors, and line straightening as described
previously. The algorithm starts with calculating the vector orthogonal to the
first two points of a line (solid arrow in
Figure~\ref{fig:algo_break_upA}). For each following point, the vector from the
starting point is calculated (dashed arrow in
Figure~\ref{fig:algo_break_upA}). Afterwards, the angle and distance
between the orthogonal and the calculated vector is computed. For an ellipse, this angle has to shrink from $90^{\circ}$ to $0^{\circ}$. Once the angle has reached $0^{\circ}$, it has to
grow back to $90^{\circ}$ whereas the distance has to shrink. If this is not
the case in the beginning, the orthogonal vector has to be turned
over.
In case the shrinking and growing do not apply to the behavior of the
line, a point where the edge has to be split is found (Figure~\ref{fig:algo_break_upB}
and \ref{fig:algo_break_upC}).
This is shown in more detail in the provided pseudocode in
Algorithm~\ref{alg:line_break}.

\subsection{Select best ellipse}
In this step, all consecutive edge pixels in the edge image are collected. For
the morphologic approach, this is done by combining all connected edge pixels into a line.
For the algorithmic approach the closed lines (all pixels in the line are connected) can be excluded to decrease runtime. Therefor the open lines (start and end pixel have only one neighbor) and the closed lines have to be separated. Open lines are collected by starting new lines only on pixels with one neighbor and closed lines are collected by starting at any pixel not accessed in the first step.
These lines are evaluated based on their shape, the
resulting shape after an ellipse fit, and the image intensity enclosed by the ellipse.
\begin{figure}[h]
	\centering
	\begin{subfigure}[b]{0.6\marginparwidth}
		\centering
		\includegraphics[width=1.0\textwidth,natwidth=384,natheight=288]{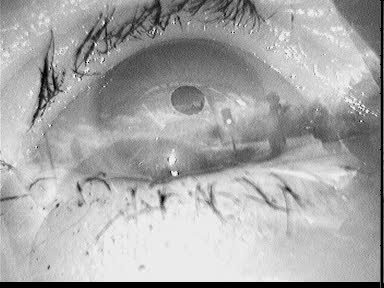}
		\caption{}
		\label{fig:ellipse_evalA}
	\end{subfigure}
	\begin{subfigure}[b]{0.6\marginparwidth}
		\centering
		\includegraphics[width=1.0\textwidth,natwidth=384,natheight=288]{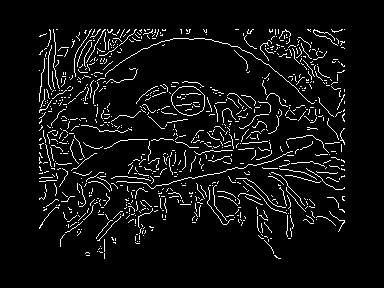}
		\caption{}
		\label{fig:ellipse_evalB}
	\end{subfigure}
	\begin{subfigure}[b]{0.6\marginparwidth}
		\centering
		\includegraphics[width=1.0\textwidth,natwidth=384,natheight=288]{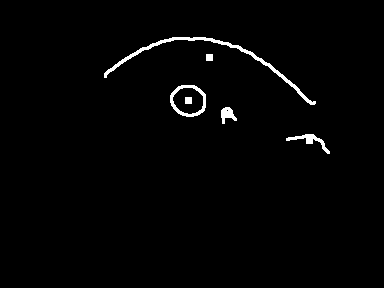}
		\caption{}
		\label{fig:ellipse_evalC}
	\end{subfigure}
	
	\begin{subfigure}[b]{0.6\marginparwidth}
		\centering
		\includegraphics[width=1.0\textwidth,natwidth=384,natheight=288]{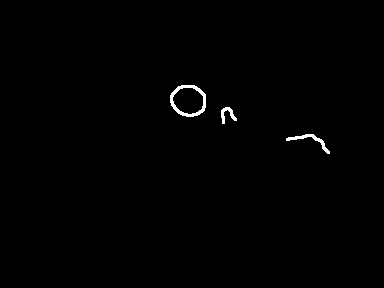}
		\caption{}
		\label{fig:ellipse_evalD}
	\end{subfigure}
	\begin{subfigure}[b]{0.6\marginparwidth}
		\centering
		\includegraphics[width=1.0\textwidth,natwidth=384,natheight=288]{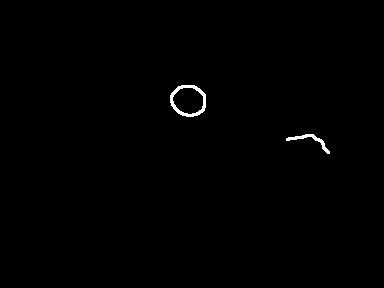}
		\caption{}
		\label{fig:ellipse_evalE}
	\end{subfigure}
	\begin{subfigure}[b]{0.6\marginparwidth}
		\centering
		\includegraphics[width=1.0\textwidth,natwidth=384,natheight=288]{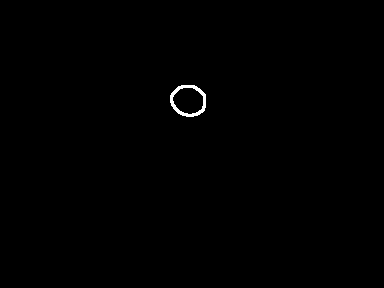}
		\caption{}
		\label{fig:ellipse_evalF}
	\end{subfigure}
	\caption{ (\subref{fig:ellipse_evalA}) input image and (\subref{fig:ellipse_evalB}) edge-filtered result. For each line, we analyze whether it is curved based on the centroid of its pixels. The result is shown
		in (\subref{fig:ellipse_evalC}) with the mean positions as bright
		dots. Then the algorithm fits an ellipse to the line. In case of success, the ellipse is further analyzed. Remaining lines after this fitting step are shown in (\subref{fig:ellipse_evalD}). The first evaluation of the ellipse filters stretched ellipses by comparing the ratio of the two ellipse radii. The result is shown in (\subref{fig:ellipse_evalE}).
		For the pupil area restriction a maximum and minimum percentage of pixels
		in the image is used as parameters. Picture (\subref{fig:ellipse_evalF}) shows the
		remaining contour after this step.}
	\label{fig:ellipse_eval}
\end{figure}
\subsubsection{Remove straight lines}
Since pupil contours exhibit a round or elliptical shape, straight lines have to be removed. For each line, we analyze whether it is straight or curved based on the mean position of all pixels belonging to it.
If the shortest distance of a line pixel to the mean position is below an
empirically set threshold $min\_mean\_line\_dist$, the line is straight. Note
that this decision is taken for both x and y dimensions. An example
result of such a step is shown in Figure~\ref{fig:ellipse_evalC} where the mean position is represented by a white dot.
\subsubsection{Ellipse fitting}
There are several ways to fit an ellipse to a set of coordinates. 
In case of an online scenario (such as driving), where the information about the pupil position is used as input to other systems (e.g. driver assistance), we are interested in very low latencies. Therefore, we employ the least squares ellipse fit as in~\cite{fitzgibbon1999direct} for efficient ellipse fitting. An exemplary result is shown in Figure~\ref{fig:ellipse_evalD}.
\subsubsection{Ellipse evaluation}
\label{subsub:ellipse-evaluation}
\begin{figure}[h]
	\centering
	\begin{subfigure}[b]{0.35\marginparwidth}
		\centering
		\includegraphics[width=1.0\textwidth,natwidth=43,natheight=51]{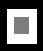}
		\caption{}
		\label{fig:ellipse_eval_patternA}
	\end{subfigure}
	\begin{subfigure}[b]{0.415\marginparwidth}
		\centering
		\includegraphics[width=1.0\textwidth,natwidth=60,natheight=60]{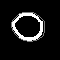}
		\caption{}
		\label{fig:ellipse_eval_patternB}
	\end{subfigure}
	\caption{Calculation of the difference between the inner and outer area of an ellipse.
	}
	\label{fig:ellipse_eval_pattern}
\end{figure}
In this step, we exclude ellipses that are unlikely to describe the pupil by
considering their area, shape, and gray value ratio between the area inside and
outside of the ellipse. The first restriction pertains the shape of the pupil by
restricting the ratio between the two ellipse radii. The rationale is that the
pupil position relative to the eye tracker camera can only distort the pupil ellipse eccentricity to a certain point. In our implementation, we
chose $radi_{ratio}=3$ empirically.
The second restriction regards the pupil
area in relation to the image size, since the eye tracker camera has to be
positioned at a restricted distance from the eye (neither to close nor to far),
which is reflected on the ratio of the image area occupied by the pupil.
We used two thresholds, namely $min_{area}=0.5\%$ and $max_{area}=10\%$ of the
total image area.
Due to the eye physiology, the last evaluation step expects the pupil to
be darker than its surroundings.
Figure~\ref{fig:ellipse_eval_patternA} shows the calculated pattern based on the
radius of the ellipse. To save computation time, instead of the whole ellipse is, we consider only a portion of the minimum enclosing, unrotated rectangle,
as shown in Figure~\ref{fig:ellipse_eval_patternA}. Pixels within the gray box
in Figure~\ref{fig:ellipse_eval_patternA} contribute to the pupil intensity
value and those within the black box contribute to the surrounding intensity.
The size of the gray box is $\frac{1}{2}$ of the width and height of the
enclosing rectangle. The white box in Figure~\ref{fig:ellipse_eval_patternA} has
the size of the enclosing rectangle and the surrounding black box has
$\frac{3}{2}$ of this size.\\
To evaluate the validity of an ellipse the surface difference of the pupil box and the surrounding box as shown in Figure~\ref{fig:ellipse_eval_patternA} is calculated. This difference value is
compared against a threshold. In our implementation, we used a $validity_
{threshold}=10$ implying that we expect the surface difference to exceed 10 in order to be valid.
\subsubsection{Rate ellipse}
All found ellipses have to be compared against each other. For this, the inner
gray value of each ellipse is computed by calculating a vector between each
point of the line and the center of the ellipse. This vector is shortened by
multiplying it stepwise from 0.95 to 0.80 with a step size of 0.01. Figure~\ref{fig:ellipse_eval_patternB} shows the line
pixels in gray and all pixels contributing to the inner gray value
($gray_{value}$ in Equation~(\ref{eq:eval})) in white.
Note that each pixel can only contribute once to the
inner gray value. This value is normalized by the sum of all contributing pixels.
\begin{equation}
	eval(el)=gray_{value}*(1+\lvert el_{width}-el_{height}\rvert)
	\label{eq:eval}
\end{equation}
The best of all remaining ellipses is chosen by selecting the ellipse with the
lowest inner gray value and the roundest shape. Equation~(\ref{eq:eval}) shows the formula
for calculating the rank of an ellipse, where $el$ is the ellipse, and $el_{width},
el_{height}$ are the radii of the ellipse. If $el_{width}$ and
$el_{height}$ are equal the ellipse is round. The variable $gray_{value}$ in
Equation~(\ref{eq:eval}) is the calculated inner gray value as specified before. The ellipse
with the lowest value calculated based on Equation~(\ref{eq:eval}) is chosen. If there is
more than one ellipse with the lowest value, the one with the most edge points and therefore clearest contour is chosen.
\subsection{Coarse positioning}
If the algorithm cannot find a good pupil edge, e.g. due to motion blur,  pupil being located in a dark spot, or when the pupil is hidden behind eyelashes, a different approach is chosen. More specifically, we apply an additional method that tries to find the pupil by first determining a likely location candidate and then refining this position. Since a computationally demanding convolution operation is required, we rescale the image to keep
run-time tractable (see Section~\ref{subsec:rescaleimg}). This rescaling process
contains a low pass procedure to preserve dark regions and to reduce the effect
of blurring or eyelashes.
Afterwards, the image is convolved with two different filters separately: 1) a
surface difference filter to  calculate the area difference between an inner
circle and a surrounding box, and 2) a mean filter. The results of both
convolutions are multiplied, and the maximum value is set as
the starting point of the refinement step.
\subsubsection{Rescale image with low pass}
\label{subsec:rescaleimg}
\begin{figure}[h]
	\centering
	\begin{subfigure}[b]{0.6\marginparwidth}
		\centering
		\includegraphics[width=1.0\textwidth,natwidth=384,natheight=288]{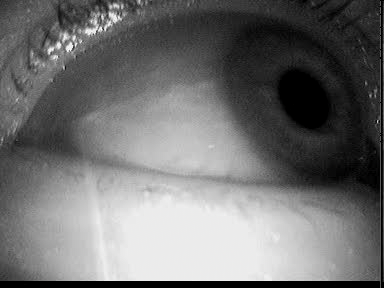}
		\caption{}
		\label{fig:mum_resA}
	\end{subfigure}
	\begin{subfigure}[b]{0.6\marginparwidth}
		\centering
		\includegraphics[width=1.0\textwidth,natwidth=620,natheight=460]{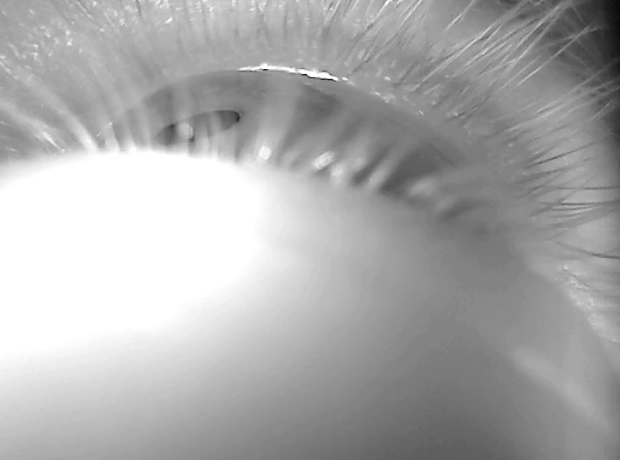}
		\caption{}
		\label{fig:mum_resB}
	\end{subfigure}
	\begin{subfigure}[b]{0.6\marginparwidth}
		\centering
		\includegraphics[width=1.0\textwidth,natwidth=384,natheight=288]{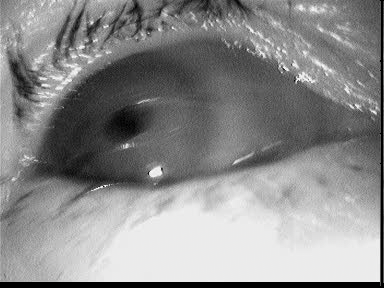}
		\caption{}
		\label{fig:mum_resC}
	\end{subfigure}
	
	\begin{subfigure}[b]{0.6\marginparwidth}
		\centering
		\includegraphics[width=1.0\textwidth,natwidth=64,natheight=48]{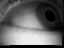}
		\caption{}
		\label{fig:mum_resD}
	\end{subfigure}
	\begin{subfigure}[b]{0.6\marginparwidth}
		\centering
		\includegraphics[width=1.0\textwidth,natwidth=103,natheight=76]{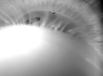}
		\caption{}
		\label{fig:mum_resE}
	\end{subfigure}
	\begin{subfigure}[b]{0.6\marginparwidth}
		\centering
		\includegraphics[width=1.0\textwidth,natwidth=64,natheight=48]{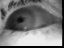}
		\caption{}
		\label{fig:mum_resF}
	\end{subfigure}
	\caption{(\subref{fig:mum_resA}), (\subref{fig:mum_resB}) and
		(\subref{fig:mum_resC}) show input images taken from the
		data set proposed by~\protect\cite{fuhl2015excuse} and
		~\protect\cite{swirski2012robust}. The results of the downscaling operation by a factor of six using the mean between zero and the mean of the input image region 		influencing a pixel are shown in (\subref{fig:mum_resD}), (\subref{fig:mum_resE}) and
		(\subref{fig:mum_resF}), respectively.
	}
	\label{fig:mum_res}
\end{figure}
There are several methods to downscale an image, e.g. based on nearest neighbor,
bilinear or bicubic interpolations, based on Lanczos kernel or more advanced downscaling operations like content adaptive~\cite{kopf2013content} or clustering based~\cite{gerstner2012pixelated} downscaling. 

\begin{figure}[h]
	\centering
	\begin{subfigure}[b]{0.8\marginparwidth}
		\centering
		\includegraphics[width=1.0\textwidth,natwidth=464,natheight=346]{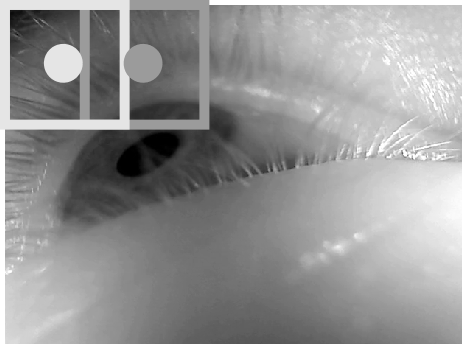}
		\caption{}
		\label{fig:mum_calcA}
	\end{subfigure}
	\begin{subfigure}[b]{0.8\marginparwidth}
		\centering
		\includegraphics[width=1.0\textwidth,natwidth=510,natheight=356]{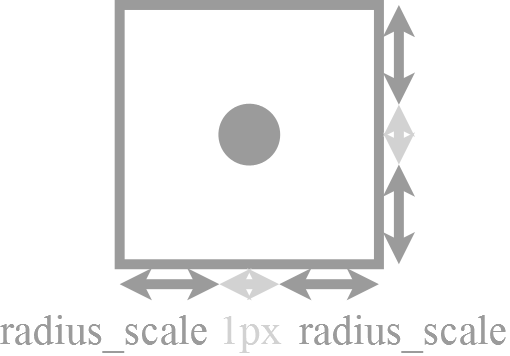}
		\caption{}
		\label{fig:mum_calcB}
	\end{subfigure}
	\caption{ (\subref{fig:mum_calcA}) shows how neighborhood regions of pixels close to each other in the downscaled image can overlap (light gray
		box and dark gray box). Each gray box represents the pixels
		influencing the intensity of a pixel in the downscaled image. The circles represent the center of a region. In (\subref{fig:mum_calcB}) the construction of the window based on the parameter $radius_{scale}$ is shown.
	}
	\label{fig:mum_calc}
\end{figure}

In case the edge detection part of the algorithm could not find a good edge because of motion blur (Figure~\ref{fig:mum_resC}) or eyelashes
(Figure~\ref{fig:mum_resB}), a downscaling operation that weights dark pixels
stronger would be preferable. However, considering that the pupil could also be
in a dark region of the image (as in Figure~\ref{fig:mum_resA}), weighting dark
pixels too strong can lead to a merging of the pupil and the
surrounding dark region. We apply a fast method to calculate the intensity
histogram and the mean (Equation~(\ref{eq:eq2})) of the pixels influencing the
new pixel. Afterwards, the mean of the lower part of the histogram (defined as
the part smaller than the mean of the whole histogram) is computed
(Equation~(\ref{eq:eq3})). The resulting value is used as the intensity of the
new pixel. This method weights dark pixels stronger based on the intensity distribution of the influencing area.
\begin{equation}
	Mean(x_1,y_1,x_2,y_2)=\frac{\sum_{x_i=x_1}^{x_2}\sum_{y_i=y_1}^{y_2} I(x_i,y_i)}{\lvert x_1-x_2\rvert*\lvert y_1-y_2\rvert}
	\label{eq:eq2}
\end{equation}
\begin{equation}
	MUM(x_1,y_1,x_2,y_2)=\frac{\sum_{x_i=0}^{Mean(x_1,y_1,x_2,y_2)} IH(x_i)*x_i}{\sum_{x_i=0}^{Mean(x_1,y_1,x_2,y_2)} IH(x_i)}
	\label{eq:eq3}
\end{equation}
where $x_1, y_1, x_2$ and $y_2$ are the coordinates defining the considered
neighborhood area that influences the new pixel intensity. $I(x_i,y_i)$ denotes
the intensity value of a pixel.\\
Equation~\ref{eq:eq3} yields the mean neighborhood intensity of the dark neighborhood regions, where darkness is defined by the mean calculated in Equation~(\ref{eq:eq2}). Therefore, it uses the intensity histogram of the region which is denoted as $IH(x_i)$ and the intensity index denoted as $x_i$. For our implementation we used overlapping regions with a window
$radius_{scale}=5$ (Figure~\ref{fig:mum_calcA}). The overlapping regions do not include the center of the other boxes, and therefore, $radius_{scale}=5$
downscales an image by a factor of six (Figure~\ref{fig:mum_calcB}).

\subsubsection{Convolution filters}
\begin{figure}[h]
	\centering
	\begin{subfigure}[b]{0.5\marginparwidth}
		\centering
		\includegraphics[width=1.0\textwidth,natwidth=17,natheight=17]{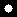}
		\caption{}
		\label{fig:conv_filterA}
	\end{subfigure}
	\begin{subfigure}[b]{0.5\marginparwidth}
		\centering
		\includegraphics[width=1.0\textwidth,natwidth=17,natheight=17]{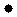}
		\caption{}
		\label{fig:conv_filterB}
	\end{subfigure}
	\begin{subfigure}[b]{0.5\marginparwidth}
		\centering
		\includegraphics[width=1.0\textwidth,natwidth=441,natheight=476]{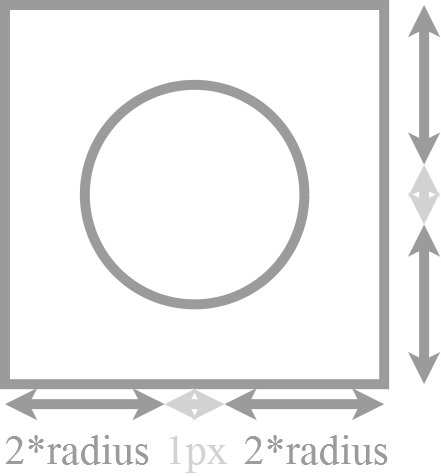}
		\caption{}
		\label{fig:conv_filterC}
	\end{subfigure}
	\caption{ (\subref{fig:conv_filterA}) mean filter, where the white region's sum is one and the black region's is zero. (\subref{fig:conv_filterB}) surface difference filter, where the black inner circle sums up to minus one and the surrounding white to one. Both kernels have the same size. (\subref{fig:conv_filterC}) filter construction, where the radius is calculated based on the image resolution.
	}
	\label{fig:conv_filter}
\end{figure}
The convolution filters used are a mean (Figure~\ref{fig:conv_filterA})
and a surface difference filter (Figure~\ref{fig:conv_filterB}). Because of the
unknown shape and expected roundness of the pupil both filters contain the shape
of a circle. Our algorithm expects the input image to contain the complete eye, and
therefore, the expected pupil size depends on image resolution. To calculate the parameter $radius_{filter}$ we simply divide the resolution in the \emph{x} and \emph{}y dimension of
the image by 100. Afterwards, the maximum of these two values is rounded up and
used as the parameter $radius_{filter}$. The construction of the filters based
on this value ($radius_{filter}=radius$) is shown in
Figure~\ref{fig:conv_filterC}. \\ The diameter of such a circle in the real
image is
\begin{equation*}
	(radius_{scale}+1)*(radius_{filter}*2+1),
\end{equation*}
which is expected to be
larger than the real pupil. This is important for the surface difference filter
(Figure~\ref{fig:conv_filterB}) because on larger pupils the result in the
middle would be lower than the result closer to the border of the pupil.
\subsubsection{Select best position}
\begin{figure*}[!ht]
	\centering
	\includegraphics[width=1.0\textwidth]{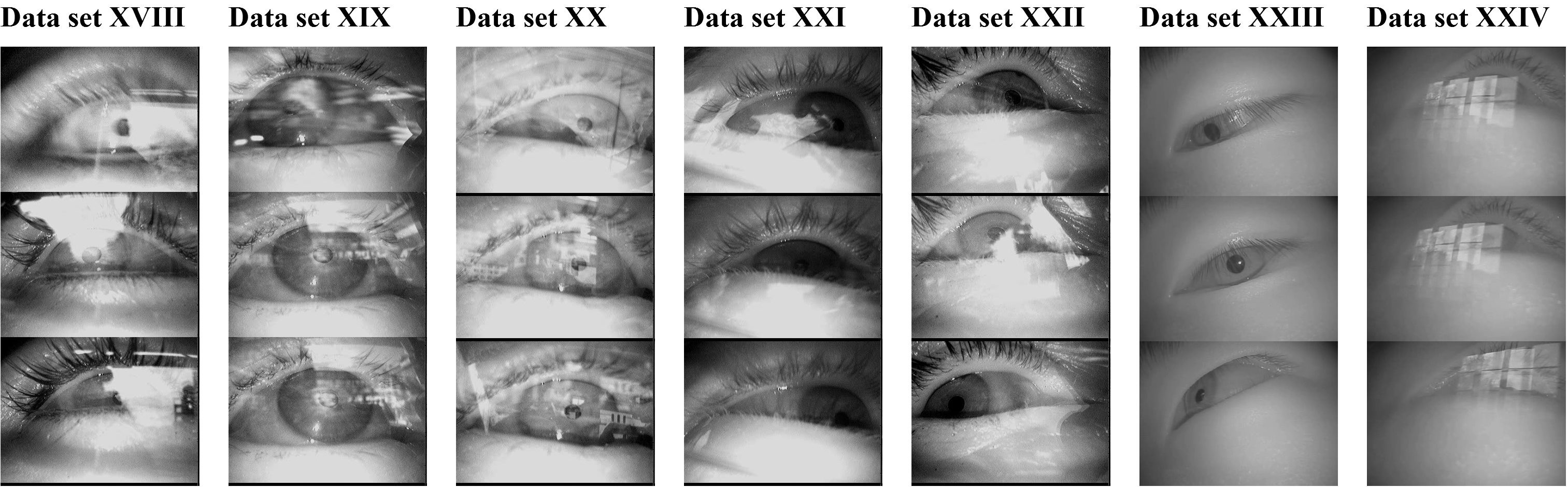}
	\caption{New hand-labeled data sets with exemplary images showing the main challenges regarding the automated image processing. The first five data sets were collected in an on road driving experiment, the last two are collected during in-door experiments with Asian subjects.}
	\label{fig:new_datasets}
\end{figure*}
\begin{figure}[h]
	\centering
	\includegraphics[width=0.3\textwidth]{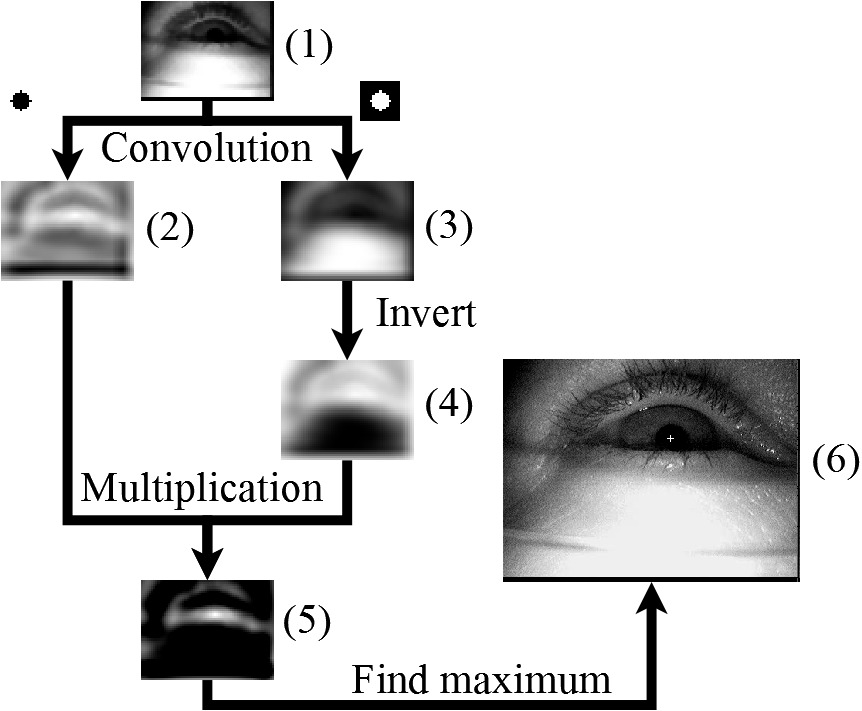}
	\caption{Workflow after downscaling of the coarse positioning. The input (1) is the downscaled image. (2) result of the convolution with the surface difference filter (Figure~\ref{fig:conv_filterA}). (3) convolution result of the mean filter (Figure~\ref{fig:conv_filterB}) and (4) 		inverted image. The result (5) is the point wise multiplication of (2)
		and (4). The absolute maximum of (5) is represented by a white cross in the real image (6) taken from the data set proposed in~\protect\cite{fuhl2015excuse}.
	}
	\label{fig:select_best}
\end{figure}
To find the best fitting position of the pupil we first convolve the downscaled
image with the surface difference filter  (Figure~\ref{fig:conv_filterB}). All
areas with low intensity values in the inner circle of the filter (black
in Figure~\ref{fig:conv_filterB}) and high values in the surrounding area will
have positive results (white in Figure~\ref{fig:select_best}(2)). The bigger this difference is, the higher the convolution response. The idea behind this
is that the pupil is surrounded by brighter intensity values. Problems with this filter are that other areas respond also with positive values and the filter response does not include intensity information of the inner area (black in Figure~\ref{fig:conv_filterB}). We are searching for the pupil, which is expected to be dark, and, therefore, we use the mean filter
(Figure~\ref{fig:conv_filterA}) to include the intensity response of the inner
area (Figure~\ref{fig:select_best}(3)). To achieve this, the result of the
convolution with the mean filter has to be inverted  (Figure~\ref{fig:select_best}(4)). This is because the response of areas with
low intensity is low, and, to use it as weight for the result of the surface
difference filter, we want it to be high.\\
The weighting is done by pointwise multiplication of the two convolution
responses of the inverted mean (Figure~\ref{fig:select_best}(4)) and the surface difference filter (Figure~\ref{fig:select_best}(2)). In the result of the weighting (Figure~\ref{fig:select_best}(5)) the maximum is searched and used as coarse position (white cross in Figure~\ref{fig:select_best}(6)).
\subsection{Optimize position}
\begin{figure}[h]
	\centering
	\begin{subfigure}[b]{0.75\marginparwidth}
		\centering
		\includegraphics[width=1.0\textwidth,natwidth=384,natheight=288]{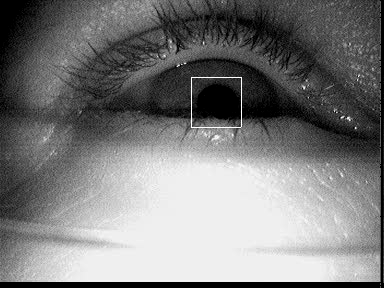}
		\caption{}
		\label{fig:opti_posA}
	\end{subfigure}
	\begin{subfigure}[b]{0.75\marginparwidth}
		\centering
		\includegraphics[width=1.0\textwidth,natwidth=384,natheight=288]{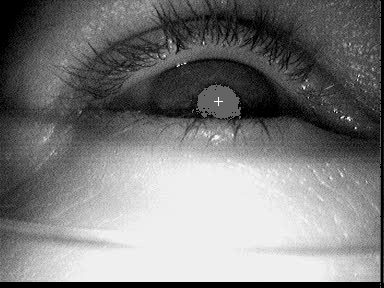}
		\caption{}
		\label{fig:opti_posB}
	\end{subfigure}
	\caption{ In (\subref{fig:opti_posA}), the area, in which the optimization takes place is enclosed by a white box. (\subref{fig:opti_posB}) shows the pixels below the calculated threshold (dark gray area) and the resulting position (white cross).
	}
	\label{fig:opti_pos}
\end{figure}
The coarse position is based on the downscaled image, and, therefore,
one pixel error relative to the pupil center represents a distance of six pixels in the
original image. For the optimization step we expect the coarse position to be contained within the pupil and calculate a
pupil intensity threshold using the neighborhood of the coarse position in the real image. In
our implementation we choose $size_ {neigbourhood}=2$, meaning a pixel
distance of two in each direction. The mean of this box is calculated, and the
absolute difference to the pixel value of the coarse position is
computed. This difference is added to the coarse position pixel value and used as a
threshold. To optimize the position, a small window (Figure~\ref{fig:opti_posA}, white box)
surrounding the coarse position in the real image is thresholded. The dark gray area in Figure~\ref{fig:opti_posB} shows this thresholded region. We chose the window size $radius_{filter}*radius_{filter}$ in each direction empirically. Afterwards, the center of mass of the thresholded pixels is calculated and used as pupil center position (white cross in Figure~\ref{fig:opti_posB}).
\subsection{Validate position}
The second method will always find a pupil location, even if the eye is currently closed.
Therefore, we have to validate the candidate location. This is done in the
same way as for the ellipse validation shown in
Figure~\ref{fig:ellipse_eval_patternA}. For the two diameters of the ellipse
we used $radius_{filter}*radius_{filter}*2 +1$.
The parameter $validity_ {threshold}$ is set to the value previously defined in
Section~\ref{subsub:ellipse-evaluation} (i.e., 10).

\section{Experimental Evaluation}
ElSe was evaluated on over 94,000 hand-labeled eye images collected during eye-tracking experiments in real-world scenarios. These data sets and the performance of our algorithm will be presented in the following.
\subsection{Data Sets}
For our evaluation, we employed several data sets provided in related work as
well as a new hand-labeled data set. More specifically, ElSe was evaluated on
the data set presented by Swirski et al.~\cite{swirski2012robust}, 17 data sets
introduced by ExCuSe~\cite{fuhl2015excuse} (i.e., data sets I-XVII in
Table~\ref{tbl:data_sets}) and 7 new hand-labeled data sets (i.e., data sets
XVIII-XXIII in Table~\ref{tbl:data_sets}). Figure~\ref{fig:new_datasets} shows
an overview of these new data sets, where each column contains exemplary images
of one data set. Among these, Data sets XVIII-XXII were derived from
eye-tracking recordings during driving~\cite{kasneci2014driving}. Data sets
XXIII and XXIV were recorded during in-door experiments with Asian subjects,
where the challenge in pupil detection arises from eyelids and eyelashes
covering or casting shadows onto the pupil (and, in one case, glasses
reflections). Further challenges associated with Data set XXIV (Figure~\ref{fig:new_datasets} last column) are related to reflections on eyeglasses. These reflections have low transparency and, therefore, affect the intensity value of the pupil in the image as well as the edge image.\\
The challenges in the eye images included in the Data sets XVIII, XIX, XX, XXI
and XXII are related to motion blur, reflections, and low pupil contrast to the
surrounding area. These challenges occur simultaneously in some images. For Data
set XVIII (Figure~\ref{fig:new_datasets} first column) most of the reflections
have low transparency and form few areas. In contrast to that, reflections in
Data set XIX (Figure~\ref{fig:new_datasets} second column) also have low
transparency but appear scattered in many areas. This leads to edge images where
the pupil edges are very difficult to extract. For Data set XX, the reflections
are also scattered in many areas but with more transparency. Data set XXI
consists mainly of images where the main challenge is a dark region surrounding
the pupil. This is
shown in the fourth column of Figure~\ref{fig:new_datasets}. All seven data sets
were recorded with different subjects and contain overall 55,712 images
(resolution 384x288 pixels). These data set can be downloaded at
\textit{ftp://emmapupildata@messor.informatik.uni-tuebingen.de} (password:\textit{eyedata}).
\subsection{Results}
\setlength{\tabcolsep}{1.5mm}
\setlength{\extrarowheight}{-1px}
\begin{table}[h]
	\small
	\centering
	\rowcolors{1}{}{lightgray}
	\begin{tabular}{lccccc}
		\toprule
		\textbf{Data set} & \textbf{SET(\%)} & \textbf{Starburst(\%)} &
		\textbf{Swirski(\%)}
		& \textbf{ExCuSe(\%)} & \textbf{ElSe(\%)}\\[-4px]
		\midrule
		Swirski  & 63.0 & 19.33 & 77.17 & \textbf{\emph{86.17}} & 82 \\
		I  & 10.27 & 5.48 & 5.11 & 70.95 & \textbf{\emph{85.52}} \\
		II  & 43.76 & 4.16 & 26.34 & 34.26 & \textbf{\emph{65.35}} \\
		III  & 12.23 & 1.71 & 6.81 & 39.44 & \textbf{\emph{63.60}} \\
		IV  & 4.03 & 4.44 & 34.54 & 81.58 & \textbf{\emph{83.24}} \\
		V  & 18.08 & 14.66 & 77.85 & 77.28 & \textbf{\emph{84.87}} \\
		VI  & 10.3 & 19.14 & 19.34 & 53.18 & \textbf{\emph{77.52}} \\
		VII  & 2.19 & 2.41 & 39.35 & 46.91 & \textbf{\emph{59.51}} \\
		VIII  & 36.67 & 9.52 & 41.90 & 56.83 & \textbf{\emph{68.41}} \\
		IX  & 10.2 & 13.88 & 24.09 & 74.60 & \textbf{\emph{86.72}} \\
		X  & 57.62 & 51.07 & 29.88 & \textbf{\emph{79.76}} & 78.93 \\
		XI  & 23.51 & 27.79 & 20.31 & 56.49 & \textbf{\emph{75.27}} \\
		XII  & 56.11 & 64.50 & 71.37 & 79.20 & \textbf{\emph{79.39}} \\
		XIII  & 33.40 & 46.64 & 61.51 & 70.26 & \textbf{\emph{73.52}} \\
		XIV  & 46.27 & 22.81 & 53.3 & 57.57 & \textbf{\emph{84.22}} \\
		XV  & 38.29 & 7.71 & \textbf{\emph{60.88}} & 52.34 & 57.30 \\
		XVI  & 57.14 & 8.93 & 17.86 & 49.49 & \textbf{\emph{59.95}} \\
		XVII & \textbf{\emph{91.04}} & 0.75 & 70.9 & 77.99 & 89.55 \\
		XVIII  & 1.32 & 1.92 & 12.39 & 22.24 & \textbf{\emph{50.86}} \\
		XIX  & 4.75 & 5.25 & 9.03 & 26.45 & \textbf{\emph{33.04}} \\
		XX  & 3.2 & 3.73 & 17.93 & 52.37 & \textbf{\emph{67.9}} \\
		XXI  & 2.29 & 2.41 & 8.09 & \textbf{\emph{43.54}} & 41.47 \\
		XXII  & 1.91 & 5.91 & 1.98 & 27.93 & \textbf{\emph{48.98}} \\
		XXIII  & 55.43 & 8.03 & \textbf{\emph{96.54}} & 93.86 & 94.34 \\
		XXIV  & 0.94 & 1.87 & 44.43 & 45.21 & \textbf{\emph{52.97}} \\
		\bottomrule
	\end{tabular}
	\caption{Performance comparison of SET (best result of both parameter settings), Starburst, Swirski, ExCuSe and ElSe in terms of detection rate up to an error of five pixels. The best performance on each data set is shown in bold.}
	\label{tbl:data_sets}
\end{table}

\begin{figure*}[!ht]
	\centering
	\begin{subfigure}[b]{\columnwidth}
		\centering
		\includegraphics[width=0.95\textwidth,natwidth=1501,natheight=899]{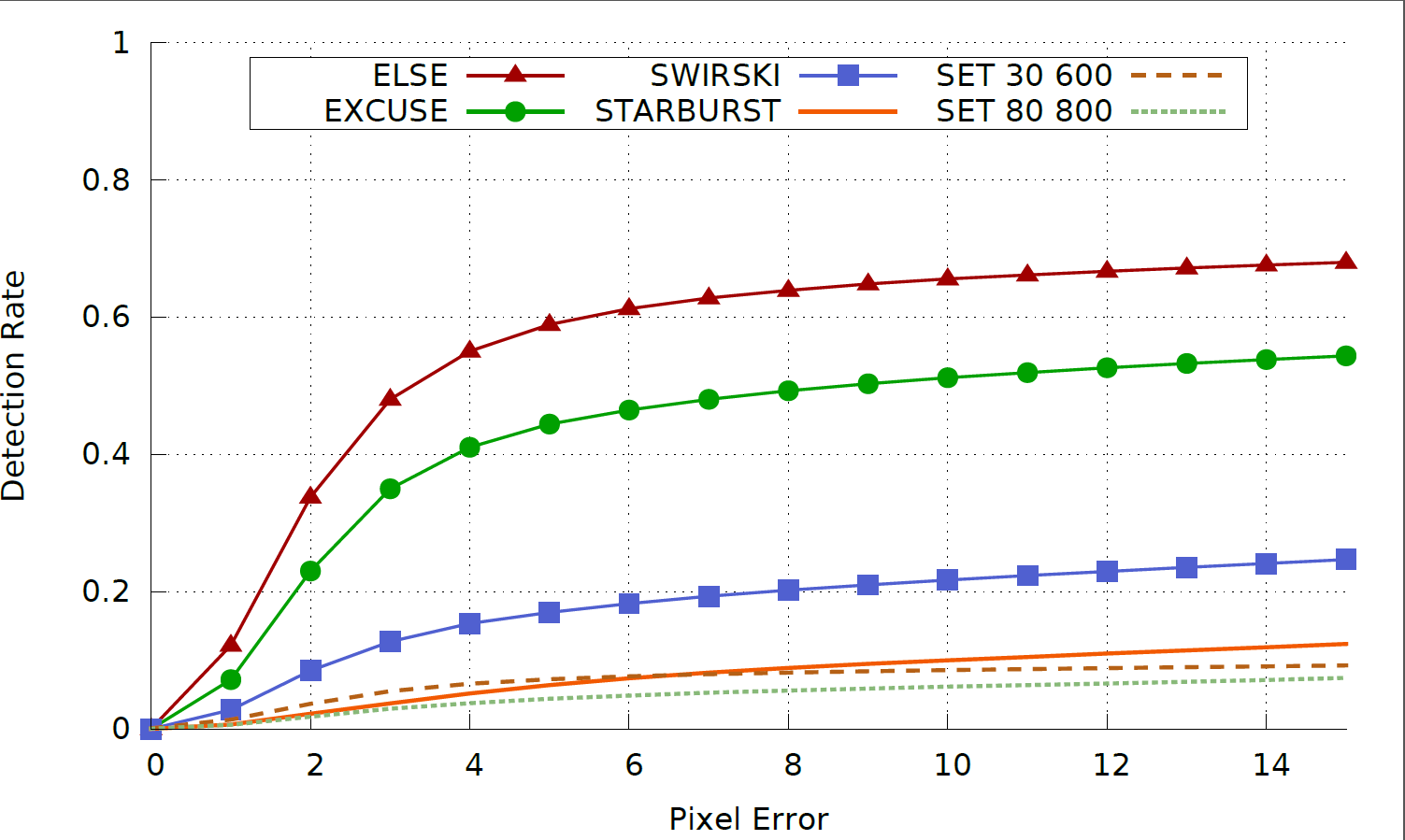}
		\caption{}
	\end{subfigure}
	\begin{subfigure}[b]{\columnwidth}
		\centering
		\includegraphics[width=0.95\textwidth,natwidth=1503,natheight=901]{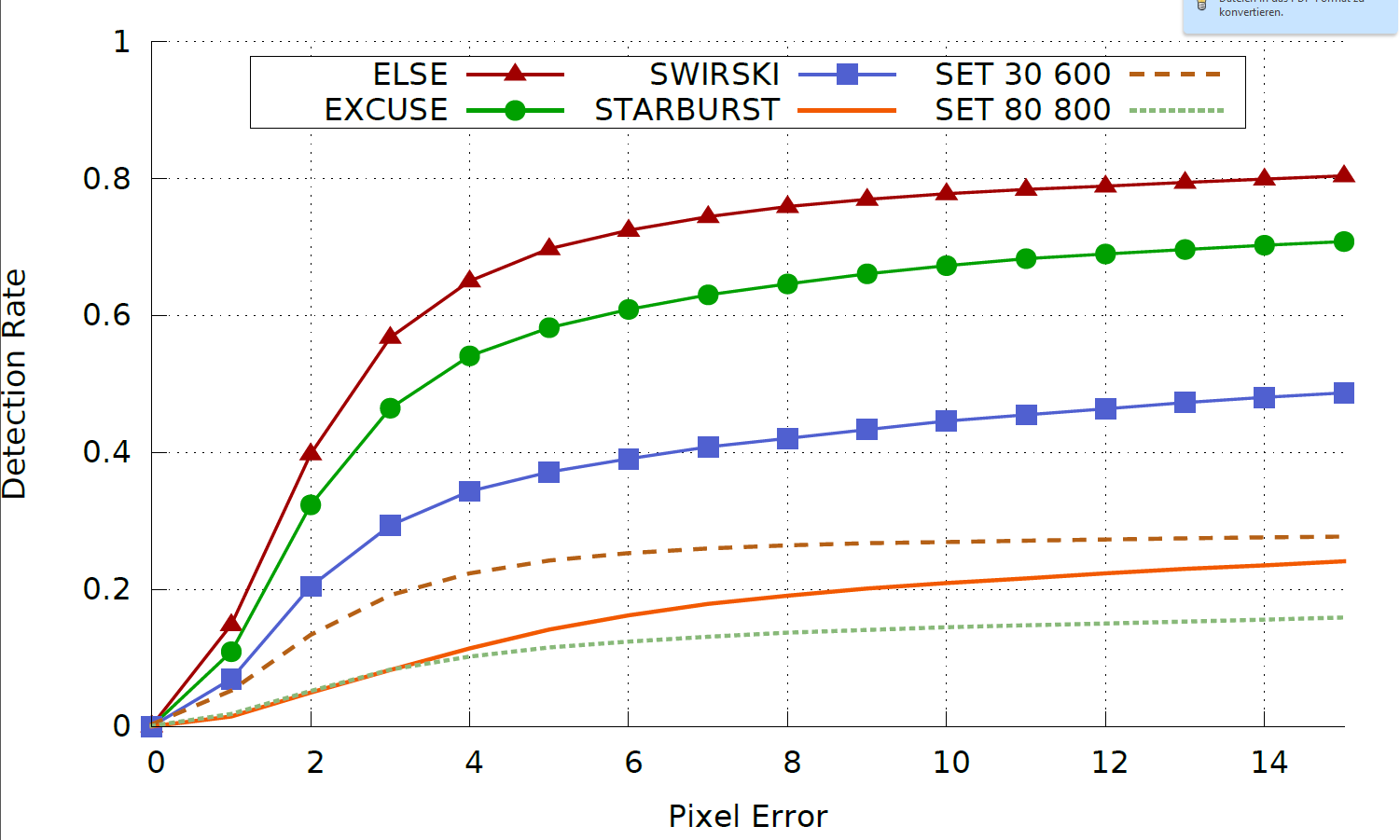}
		\caption{}
	\end{subfigure}
	\caption{
		Average detection rates at different pixel distances on all data sets.
		In (a)  the result for each data set is weighted by the number of images
		in the corresponding data set. (b) presents the mean (unweighted)
		detection rate over all data sets.
		SET~\protect\cite{javadi2015set} is shown with two different settings:
		1) luminance=30, area threshold=600 and 2) luminance=80, area
		threshold=800.
	}
	\label{fig:result_weight}
\end{figure*}

We compared our algorithm to four state-of-the-art approaches on the above data
without adjusting its parameters, i.e., ElSe was applied with one fixed
parameter setting to all data sets. The competitor algorithms are
SET~\cite{javadi2015set}, Starburst~\cite{li2005starburst}, \cite{swirski2012robust}, and ExCuSe~\cite{fuhl2015excuse}. For
SET, we applied two parameter combinations \textit{ThresholdLuminance} = 30,
\textit{ThresholdRegion} = 600 and \textit{ThresholdLuminance} = 80,
\textit{ThresholdRegion} = 800 with the MATLAB version from the eyego eyetracker
website (\textit{https://sites.google.com/site/eyegoeyetracker/}, accessed on June,
1 2015). Starburst~\cite{li2005starburst} was used in its MATLAB Version 1.1.0
as provided by the OpenEyes website
(\textit{http://thirtysixthspan.com/openEyes/software.html}, accessed on June 1,
2015) without any changes in the parameter setting. The starting location of the
algorithm was set to the center of the image. The algorithm proposed by Swirski
et al.~\cite{swirski2012robust} was used with the parameter settings provided by
the authors
(\textit{https://www.cl.cam.ac.uk/research/rainbow/projects/pupiltracking/},
accessed on June 1, 2015). ExCuSe~\cite{fuhl2015excuse} was used with the
parameter setting provided by the authors
(\textit{https://www.ti.uni-tuebingen.de/Pupil-detection.1827.0.html?\&L=1},
accessed on June 12, 2015). The performance was measured in terms of the detection rate for different pixel errors. The pixel error represents the Euclidean distance between the hand-labeled center of the pupil and the pupil center reported by the algorithm. Note that we do not report performance measures related to the gaze position on the scene, since this also depends on the calibration. We focus on the pupil center position on the eye images, where the first source of noise occurs. Table~\ref{tbl:data_sets} summarizes the performance results for each of the competing algorithms in terms of the detection rate up to an error of five pixels. For each data set, the best result is shown in bold. In addition, Figure~\ref{fig:result_weight} presents the performance of ElSe and its competitors in terms of the detection rate for different pixel error rates (0-15 pixels). More specifically, Figure~\ref{fig:result_weight}(a) shows the detection rate as the percentage of correctly detected pupil centers for the 94,713 images, whereas Figure~\ref{fig:result_weight}(b) depicts the detection rate as the average over all data sets with each data set weighted equally (due to different data set sizes).
In both evaluations, ElSe clearly outperformed all competitor algorithms. For
Data sets X, XV, XVII, XXI, XXIII, and~\cite{swirski2012robust},
ElSe has not the best detection rate but is always close to the best result (on
average 2\% worse than the best performer in this cases). \\
For the~\cite{swirski2012robust} approach, we measured a runtime of 8 ms per image on an i5-4570 (3.2GHz) CPU.
ExCuSe~\cite{fuhl2015excuse} needed 6 ms per image, whereas ElSe (without
parallelization) 7 ms. The algorithms
SET~\cite{javadi2015set} and Starburst~\cite{li2005starburst} are not comparable
in terms of runtime because we used their MATLAB version.

\subsection{Parameter sensitivity}
\begin{figure}[h]
	\centering
	\includegraphics[width=\columnwidth,natwidth=1501,natheight=901]{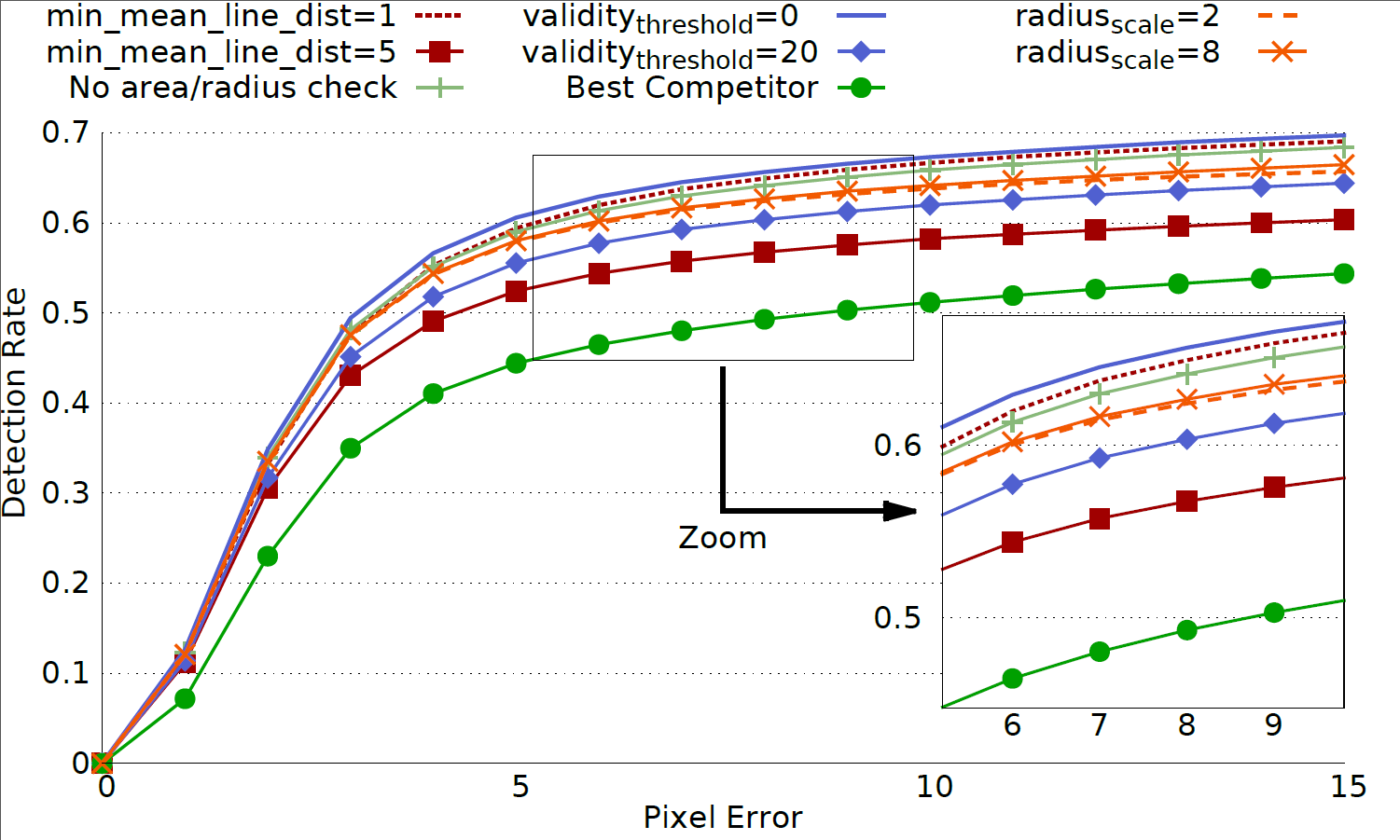}
	\caption{
		The detection rates of ElSe with different parameter settings for all
		data sets (94,713 images).
		The impact of the $min\_mean\_line\_dist$ parameter on the detection
		rate is shown in red, the impact of the $validity_ {threshold}$ in blue,
		the impact of the rescaling factor $radius_{scale}$ in orange, and the
		performance when removing the ellipse area and radius relation check in
		pale grey.
		For reference, the best competitor (ExCuSe) is also displayed.
	}
	\label{fig:impact_result}
\end{figure}
Decisions in ElSe are taken based on several parameters. Therefore, we conducted
four tests to quantify the impact of parameter settings on ElSe's performance.
These results are shown in Figure~\ref{fig:impact_result}.
The first test regards the parameter $min\_mean\_line\_dist$, which is used to
decided whether a line is straight or curved.
If the value of this parameter is too high, the probability to remove correct lines increases. A too low value does not have a significant effect on
the result but leads to an increase in runtime (1-2~ms per image).
In the second test, the ellipse restrictions for radius relation and area were
removed. This reduces the detection rate by about 1\%, and therefore, these
parameters are unlikely to have a big impact on detection results.
For the third test, the validation threshold $validity_{threshold}$ in the
algorithmic steps \textit{Select best ellipse 2.2} and \textit{Validate position
	2.5} in Figure~\ref{fig:ablauf} was changed.
For lower values of this parameter more ellipses get accepted, leading thus  to an increase in detection rate but also to a higher false positive rate. This means that more blinks are falsely accepted as pupils. Higher values decrease the
detection rate but reduce the false positive rate too.
The last test regards the coarse positioning by changing the parameter
$radius_{scale}$, which effects the rescaling. Too high and too low values for
this parameter reduce the detection rate of ElSe by about 1-2\%. The main effect
of this parameter is on the run-time because of the convolution step. For high values, the run-time decreases while the run-time increases for low values. Nonetheless, a comparison of the results in  Figure~\ref{fig:impact_result} with those shown in
Figure~\ref{fig:result_weight}~(a) reveals that even for these parameter changes ElSe has a higher detection rate than its competitors.

\subsection{Limitations of ElSe}
\begin{figure}[h]
	\centering
	\begin{subfigure}[b]{0.6\marginparwidth}
		\centering
		\includegraphics[height=0.65\textwidth,natwidth=384,natheight=288]{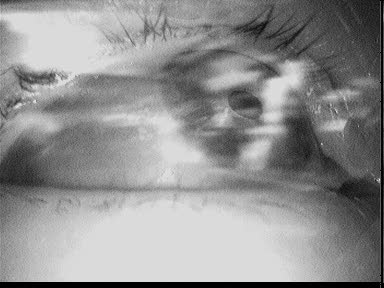}
		\caption{}
	\end{subfigure}
	\begin{subfigure}[b]{0.6\marginparwidth}
		\centering
		\includegraphics[height=0.65\textwidth,natwidth=384,natheight=288]{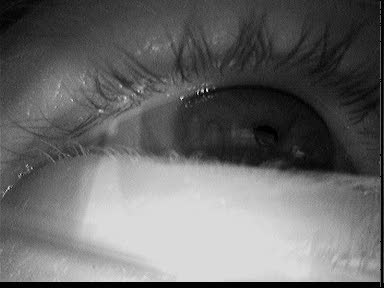}
		\caption{}
	\end{subfigure}
	\begin{subfigure}[b]{0.6\marginparwidth}
		\centering
		\includegraphics[height=0.65\textwidth,natwidth=384,natheight=288]{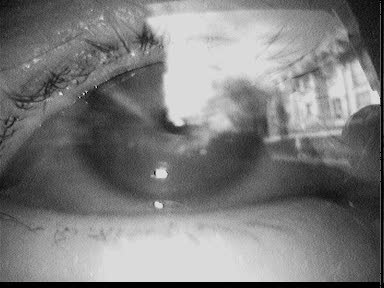}
		\caption{}
	\end{subfigure}
	
	\begin{subfigure}[b]{0.6\marginparwidth}
		\centering
		\includegraphics[height=0.75\textwidth,natwidth=100,natheight=100]{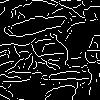}
		\caption{}
	\end{subfigure}
	\begin{subfigure}[b]{0.6\marginparwidth}
		\centering
		\includegraphics[height=0.75\textwidth,natwidth=100,natheight=100]{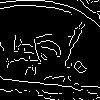}
		\caption{}
	\end{subfigure}
	\begin{subfigure}[b]{0.6\marginparwidth}
		\centering
		\includegraphics[height=0.75\textwidth,natwidth=100,natheight=100]{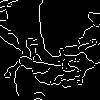}
		\caption{}
	\end{subfigure}
	
	\begin{subfigure}[b]{0.6\marginparwidth}
		\centering
		\includegraphics[height=0.65\textwidth,natwidth=64,natheight=48]{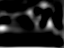}
		\caption{}
	\end{subfigure}
	\begin{subfigure}[b]{0.6\marginparwidth}
		\centering
		\includegraphics[height=0.65\textwidth,natwidth=64,natheight=48]{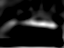}
		\caption{}
	\end{subfigure}
	\begin{subfigure}[b]{0.6\marginparwidth}
		\centering
		\includegraphics[height=0.65\textwidth,natwidth=64,natheight=48]{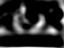}
		\caption{}
	\end{subfigure}
	\caption{Failure cases of ElSe on the input images (a), (b) and (c). (d), (e) and (f) show the morphologic filtered edge image surrounding the pupil. (g), (h) and (i) show the convolution response.}
	\label{fig:total_failure}
\end{figure}

ElSe relies on the Canny edge filter. Thus, if the edge selection fails, the convolution approach makes the critical assumption that the pupil has a low intensity value surrounded by higher intensity values. For input images where the pupil is partially covered by many small reflections (e.g., Figure~\ref{fig:total_failure} (a)) the algorithm fails. Such reflections not only destroy the edge filter response (Figure~\ref{fig:total_failure} (d)) but also lower the convolution filter result (Figure~\ref{fig:total_failure} (g)), leading thus to a wrong coarse positioning. Another example where ElSe fails, is shown in Figure~\ref{fig:total_failure}(b). Here the pupil is surrounded by a very dark iris and the skin below the eye is very bright. The dark iris leads to a good edge filter response (Figure~\ref{fig:total_failure}(e)). The edge selection, however, fails at the validation check for the correct ellipse. Afterwards, the convolution approach (Figure~\ref{fig:total_failure}(h)) is applied, but due to the bright skin below the eye, the coarse positioning fails.\\
A last failure example are reflections covering most of the pupil (Figure~\ref{fig:total_failure}(c)). In such cases the pupil edge is scattered due to the high magnitude response of the edges belonging to the reflection (Figure~\ref{fig:total_failure} (f)). Figure~\ref{fig:total_failure} (i) shows the multiplied convolution response for the coarse positioning. Here the dark pupil part (Figure~\ref{fig:total_failure} (c)) has only a low response. Since most of the pupil is bright, it will lead to a negative surface difference and, therefore, to a low weight through the mean filter.
Despite these limitations, ElSe showed high robustness in comparison to related approaches. We therefore believe that our algorithm will help to overcome obstacles related to the analysis of eye-tracking data as needed in several applications where the visual attention focus of the subject needs to be determined in an online fashion or in real-world scenarios.

\section{Conclusions}
We presented a novel pupil detection algorithm, ElSe, for real-time eye-tracking experiments
in outdoor environments. ElSe was evaluated on 94,713 challenging, hand-labeled
eye images  in which reflections, changing illumination conditions, off-axial
camera position and other sources of noise occur. We compared our approach against four state-of-the-art methods and showed that ElSe outperformed the competitors by far. The implementation of ElSe and the evaluation data sets are available for download. Thus, we highly encourage its application to outdoor eye-tracking experiments.

\bibliographystyle{acmsiggraph}
\bibliography{template4ETRA}
\end{document}